%% file: main.tex
\documentclass[12pt,a4paper]{article}

\input{macros}

\author{Jakub Sygnowski\thanks{\texttt{J.Sygnowski@students.mimuw.edu.pl}} }
\author{Henryk Michalewski\thanks{\texttt{H.Michalewski@mimuw.edu.pl}}}
\affil{University of Warsaw\\ Department of Mathematics, Informatics, and Mechanics}

\begin{document}

\title{Learning from the memory of Atari 2600}
\maketitle

\begin{abstract}
We train a number of neural networks to play games \mbox{Bowling}, Breakout and Seaquest using information stored in the memory of a video game console Atari 2600.
We consider four models of neural networks which differ in size and architecture: two networks which use only information contained in the RAM and two mixed networks which use both information in the RAM and information from the screen.

As the benchmark we used the convolutional model proposed in \cite{mnih-atari-2013}  and received comparable results in all considered games. Quite surprisingly, in the case of Seaquest we were able to train RAM-only agents which behave better than the benchmark screen-only agent. Mixing screen and RAM did not lead to an improved performance comparing to screen-only and RAM-only agents.
\end{abstract}

\input{introduction}

\input{experiment}

\input{plain}

\input{regularization}

\input{frameskip}

\input{other}

\input{conclusion}

\input{acknowledge}

\printbibliography

\newpage
\input{append}

\end{document}

%% file: macros.tex
\usepackage{array}
\usepackage{epsfig}
\usepackage[all]{xy}
\usepackage{enumerate}
\usepackage{graphicx}
\usepackage{tikz}
\usepackage{xspace}
\usepackage{float}
\usepackage[all]{xy}
\usepackage{stmaryrd}
\usepackage{amssymb}
\usepackage{amsmath}
\usepackage{verbatim}

\usepackage[backend=bibtex]{biblatex}
\bibliography{main.bib}

\usepackage{tikz}
\usepackage{xifthen}
\usepackage{intcalc}
\usetikzlibrary{calc}
\usetikzlibrary{arrows}
\usepackage{pgfplots}
\usepackage{wrapfig}
\usepackage[font={scriptsize,it}]{caption}
\usepackage{cutwin}
\usepackage{authblk}

\usepackage{enumitem, hyperref}
\makeatletter
\def\namedlabel#1#2{\begingroup
    #2%
    \def\@currentlabel{#2}%
    \phantomsection\label{#1}\endgroup
}

\usepackage{mathtools} 

\usepackage[english]{babel}
\usepackage[utf8]{inputenc}
\usepackage{graphicx}
\usepackage{amsmath}
\usepackage[colorinlistoftodos]{todonotes}

\numberwithin{equation}{section}

\renewcommand{\comment}[1]{}

\newcommand{\ignore}[1]{}

\usepackage{algorithm2e}
\SetAlFnt{\small\sf}
\SetAlgorithmName{Neural network}{List of neural networks}

\usepackage{comment}
\usepackage{subfigure}
\usepackage{appendix}

\usepackage{tabularx,colortbl}  
\usepackage{ragged2e}  
\newcolumntype{Y}{>{\RaggedRight\arraybackslash}X} 
\usepackage{booktabs}  

%% file: introduction.tex
\begin{section}{Introduction}

An Atari 2600 controller can perform one of $18$ actions\footnote{For some games only some of these 18 actions are used in the gameplay. The number of available actions is $4$ for Breakout, $18$ for Seaquest and $6$ for Bowling.} and in this work we are intended to learn which of these $18$ actions is the most profitable given a state of the screen or memory. Our work is based on deep Q-learning  \cite{mnih-atari-2013} -- a reinforcement learning algorithm that can learn to play Atari games using only  input from the screen. The deep Q-learning algorithm builds on the Q-learning \cite{qlearning} algorithm, which in its simplest form (see \cite[Figure 21.8]{norvig}) iteratively learns values $Q(state,action)$ for \emph{all} state-action pairs and lets the agent choose the action with the highest value. In the instance of Atari 2600 games this implies evaluating all pairs $(screen,action)$ where $action$ is one of the $18$ positions of the controller. This task is infeasible and similarly, learning \emph{all} values $Q(state,action)$ is not a realistic task in other real-world  games such as chess or Go.

This feasibility issues led to generalizations of the Q-learning algorithm which are intended to limit the number of parameters on which the function $Q$ depends. One can arbitrarily restrict the number of features which can be learned\footnote{E.g. we may declare that $Q(state,action) = \theta_1 f_1(state,action) + \theta_2 f_2(state,action)$, where $f_1,f_2$ are some fixed pre-defined functions, for example $f_1$ may declare value $1$ to the state-action pair $(screen,fire)$ if a certain shape appears in the bottom-left corner of the screen and $0$ otherwise and $f_2$ may declare value $1$ to $(screen,left)$ if an enemy appeared on the right and $0$ otherwise. Then the $Q$-learning algorithm learns the best values of $\theta_1,\theta_2$.}, but instead of using manually devised features, the deep Q-learning algorithm\footnote{This algorithm is also called a deep q-network or DQN.} presented in \cite{mnih-atari-2013} builds them in the process of training of the neural network. Since every neural network is a composition of a priori unknown linear maps and fixed non-linear maps, the aim of the deep Q-learning algorithm is to learn coefficients of the unknown linear maps.

In the deep Q-learning algorithm the game states, actions and immediate rewards are passed to a {\em deep convolutional network}. This type of network abstracts features of the screen, so that various patterns on the screen can be identified as similar. The network has a number of output nodes -- one for each possible action -- and it predicts the cumulative game rewards after making moves corresponding to actions.

A number of decisions was made in the process of designing of the deep Q-learning algorithm (see \cite[Algorithm 1]{mnih-atari-2013} for more details): 
(1) in each step there is some probability $\varepsilon$ of making a random move and it decreases from $\varepsilon = 1$ to $\varepsilon = 0.1$ in the course of training, (2) the previous game states are stored in the {\em replay memory}; the updates in the Q-learning are limited to a random batch of events polled from that memory, (3) the updates of unknown linear maps in the neural network are performed according to the gradient of the squared loss function which measures discrepancy between the estimation given by the network and the actual reward. In this work we use the same computational infrastructure as in \cite{mnih-atari-2013}, including the above decisions (1)-(3).

\subsection*{Related work}
The original algorithm in \cite{mnih-atari-2013} was improved in a number of ways in \cite{van2015deep,mnih-dqn-2015,liang2015state}. This includes changing network architecture, choosing better hyperparameters and improving the speed of algorithm which optimizes neural network's loss function. These attempts proved to be successful and made the deep Q-learning algorithm the state-of-the-art method for playing Atari games.

Instead of the screen one can treat the RAM state of the Atari machine as the game state. The work \cite{nir} implemented a classical planning algorithm on the RAM state. Since the Atari 2600 RAM consists of only $128$ bytes, one can efficiently search in this low-dimensional state space. Nevertheless, the learning in \cite{nir} happens during the gameplay, so it depends on the time limit for a single move. In contrast, in \cite{mnih-atari-2013} the learning process happens before the gameplay - in particular the agent can play in the real-time. To the best of our knowledge the only RAM-based agent not depending on search was presented in \cite{bellemare13arcade}. We cite these results as \texttt{ale\_ram}.

In our work we use the deep Q-learning algorithm, but instead of using screens as inputs to the network, we pass the RAM state or the RAM state and the screen together. In the following sections we describe the games we used for evaluation, as well as the network architectures we tried and hyperparameters we tweaked.

\subsection*{The work \cite{mnih-atari-2013} as the main benchmark}
The changes to the deep Q-learning algorithm proposed in \cite{mnih-dqn-2015} came at a cost of making computations more involved comparing to \cite{mnih-atari-2013}. In this work we decided to use as the reference result only the basic work \cite{mnih-atari-2013}, which is not the state of the art, but a single training of a neural network can be contained in roughly $48$ hours using the experimental setup we describe below. This decision was also motivated by a preliminary character of our study -- we wanted to make sure that indeed the console memory contains useful data which can be extracted  during the training process using the deep Q-learning algorithm. From this perspective the basic results in \cite{mnih-atari-2013} seem to be a perfect benchmark to verify feasibility of learning from RAM. We refer to this benchmark architecture as \texttt{nips} through this paper.
\end{section}

%% file: experiment.tex
\begin{section}{The setting of the experiment}

\begin{subsection}{Games}
\begin{description}
\item[Bowling] -- simulation of the game of bowling; the player aims the ball toward the pins and then steers the ball; the aim is to hit the pins \cite{bowling,bowling_man}.
\item[Breakout] -- the player bounces the ball with the paddle towards the layer of bricks; the task is to destroy all bricks; a brick is destroyed when the ball hits it \cite{breakout,breakout_man}.
\item[Seaquest] -- the player commands a submarine, which can shoot enemies and rescue divers by bringing them above the water-level; the player dies if he fails to get a diver up before the air level of submarine vanishes \cite{seaquest,seaquest_man}.
\end{description}

\begin{figure}
\begin{center}
\begin{subfigure}
  \centering
  \includegraphics[width=.325\linewidth]{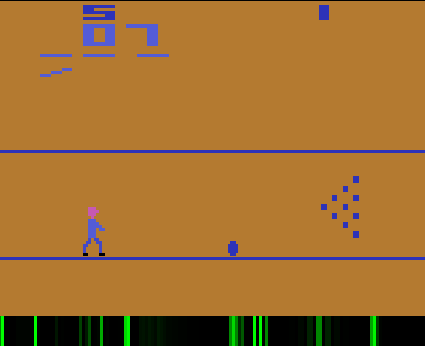}
  \label{fig:sub1}
\end{subfigure}
\begin{subfigure}
  \centering
   \includegraphics[width=.31\linewidth]{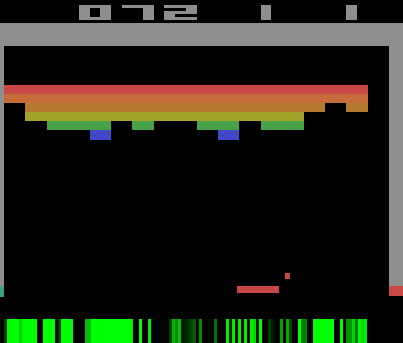}
  \label{fig:sub2}
\end{subfigure}%
\begin{subfigure}
  \centering
  \includegraphics[width=.30\linewidth]{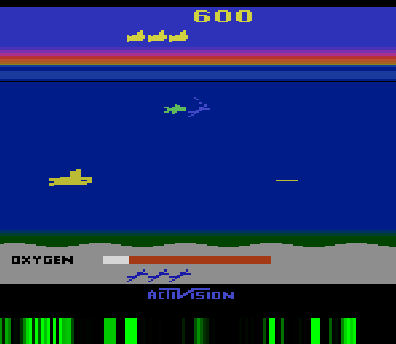}
  \label{fig:sub3}
\end{subfigure}
\caption{From left to right Bowling, Breakout and Seaquest. The 128 vertical bars and the bottom of every screenshot represent the state of the memory, with black representing $0$ and lighter color corresponding to higher values of a given memory cell. }
\end{center}
\label{fig:screenshots}
\end{figure}

We've chosen these games, because each of them offers a distinct challenge.
Breakout is a relatively easy game with player's actions limited to moves along the horizontal axis. We picked Breakout because disastrous results of learning would indicate a fundamental problem with the RAM learning.
The deep Q-network for Seaquest constructed in \cite{mnih-atari-2013} plays at an amateur human level and for this reason we consider this game as a tempting target for improvements. Also the game state has some elements that possibly can be detected by the RAM-only network (e.g. oxygen-level meter or the number of picked divers). Bowling seems to be a hard game for all deep Q-network models. It is an interesting target for the RAM-based networks, because visualizations suggest that the state of the RAM is changing only very slightly.
\end{subsection}

\begin{subsection}{Technical architecture}
By one experiment we mean a complete training of a single deep Q-network. In this paper we quote numerical outcomes of $30$ experiments which we performed\footnote{The total number of experiments exceeded $100$, but this includes experiments involving other models and repetitions of experiments described in this paper.}. For our experiments we made use of Nathan Sprague's implementation of the deep Q-learning algorithm \cite{sprague} in Theano \cite{theano} and Lasagne \cite{lasagne}. The code uses the Arcade Learning Environment \cite{bellemare13arcade} -- the standard framework for evaluating agents playing Atari games. Our code with  instructions how to run it can be found on github \cite{our-dqn}. All experiments were performed on a Linux machine equipped with Nvidia GTX 480 graphics card. Each of the experiments lasted for $1-3$ days. A single epoch of a RAM-only training lasted approximately half of the time of the screen-only training for an architecture with the same number of layers.
\end{subsection}

\begin{subsection}{Network architectures}
We performed experiments with four neural network architectures which accept the RAM state as (a  part of) the input. The RAM input is scaled by $256$, so all the inputs are between $0$ and $1$.

All the hyperparameters of the network we consider are the same as in \cite{mnih-atari-2013}, if not mentioned otherwise (see Appendix \ref{app:AppendixA}). We only changed the size of the replay memory to $\approx 10^5$ items, so that it fits into $1.5$GB of Nvidia GTX 480 memory\footnote{We have not observed a statistically significant change in results when switching between replay memory size of $10^5$ and $5\cdot 10^5$.}.
\end{subsection}
\end{section}

%% file: plain.tex
\begin{section}{Plain approach}
Here we present the results of training the RAM-only networks \texttt{just\_ram} and \texttt{big\_ram} as well as the benchmark model \texttt{nips}.

\RestyleAlgo{ruled}
\begin{algorithm}[H]
\AlFnt{\small\sf}
\caption{\texttt{just\_ram}(outputDim)}

\SetAlgoVlined
\LinesNumbered

\AlFnt{\small\sf}
\KwIn{RAM} 
\KwOut{A vector of length outputDim}
\vspace{0.05cm}

$hiddenLayer1 \leftarrow DenseLayer(RAM,128,rectify)$

$hiddenLayer2 \leftarrow DenseLayer(hiddenLayer1,128,rectify)$

$output \leftarrow DenseLayer(hiddenLayer2,outputDim,no\; activation)$

\Return{$output$}
\end{algorithm}
The next considered architecture consists of the above network with two additional dense layers:  

\RestyleAlgo{ruled}
\begin{algorithm}[H]
\AlFnt{\small\sf}
\caption{\texttt{big\_ram}(outputDim)}

\SetAlgoVlined
\LinesNumbered

\KwIn{RAM} 
\KwOut{A vector of length outputDim}
\vspace{0.05cm}

$hiddenLayer1 \leftarrow DenseLayer(RAM,128,rectify)$

$hiddenLayer2 \leftarrow DenseLayer(hiddenLayer1,128,rectify)$

$hiddenLayer3 \leftarrow DenseLayer(hiddenLayer2,128,rectify)$

$hiddenLayer4 \leftarrow DenseLayer(hiddenLayer3,128,rectify)$

$output \leftarrow DenseLayer(hiddenLayer4,outputDim,no\; activation)$

\Return{$output$}
\end{algorithm}
The training process consists of \textit{epochs}, which are interleaved by test periods. During a test period we run the model with the current parameters, the probability of doing a random action $\varepsilon = 0.05$ and the number of test steps (frames) being $10\, 000$. Figures \ref{fig:breakout-plain}, \ref{fig:seaquest-plain} and \ref{fig:bowling-plain} show the average result per \textit{episode} (full game, until player's death) for each epoch.

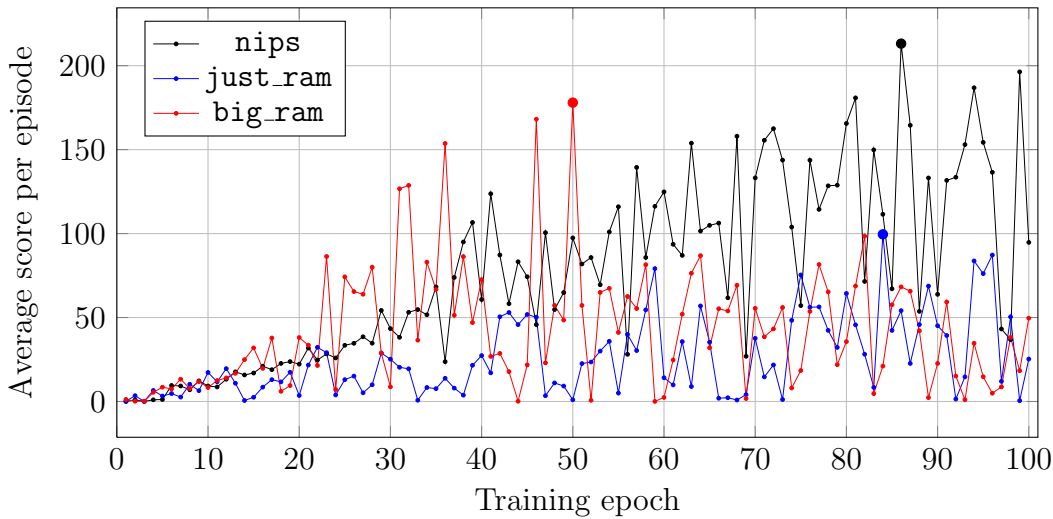
\begin{figure}[H]
\begin{center}
\hspace*{-1.5cm}\begin{tikzpicture}[scale=1]
\begin{axis}[
   xlabel={Training epoch},
   ylabel={Average score per episode},
   xmin=0,xmax=101,
   x=0.12cm,
   grid=major,
   mark options={solid, scale=0.35},
   legend entries={\texttt{nips}, \texttt{just\_ram}, \texttt{big\_ram}},
   legend style={legend pos=north west},
]
\addplot[mark=*, color=black, line width=0.25pt] table {experiments/breakout_nips.txt};
\addplot[mark=*, color=blue, line width=0.25pt] table {experiments/breakout_just_ram_100.txt};
h\addplot[mark=*, color=red, line width=0.25pt] table {experiments/breakout_big_ram_100.txt};
\addplot[mark=*, color=black, only marks, mark size=5] table {
x y
86 213.1428571
};
\addplot[mark=*, color=blue, only marks, mark size=5] table {
x	y
84	99.55555556
};
\addplot[mark=*, color=red, only marks, mark size=5] table {
x	y
50	178
};
\end{axis}
\end{tikzpicture}
\end{center}
\caption{Training results for Breakout for three plain models: \texttt{nips}, \texttt{just\_ram}, \texttt{big\_ram}. }
\label{fig:breakout-plain}
\end{figure}

\begin{figure}[htbp]
\begin{center}
\hspace*{-1.5cm}\begin{tikzpicture}[scale=1]
\begin{axis}[
   xlabel={Training epoch},
   ylabel={Average score per episode},
   xmin=0,xmax=101,
   x=0.12cm,
   grid=major,
   mark options={solid, scale=0.35},
   legend entries={\texttt{nips}, \texttt{just\_ram}, \texttt{big\_ram}},
   legend style={legend pos=north west},
]
\addplot[mark=*, color=black, line width=0.25pt] table {experiments/seaquest_nips.txt};
\addplot[mark=*, color=blue, line width=0.25pt] table {experiments/seaquest_just_ram_100.txt};
\addplot[mark=*, color=red, line width=0.25pt] table {experiments/seaquest_big_ram_100.txt};
\addplot[mark=*,color=black, only marks,mark size=5] table {
x y
82 1808
};
\addplot[mark=*, color=blue, only marks, mark size=5] table {
x y
90	1360
};
\addplot[mark=*, color=red, only marks, mark size=5] table {
x y
88	2680
};
\end{axis}
\end{tikzpicture}
\end{center}
\caption{Training results for Seaquest for three plain models: \texttt{nips}, \texttt{just\_ram}, \texttt{big\_ram}.}
\label{fig:seaquest-plain}
\end{figure}
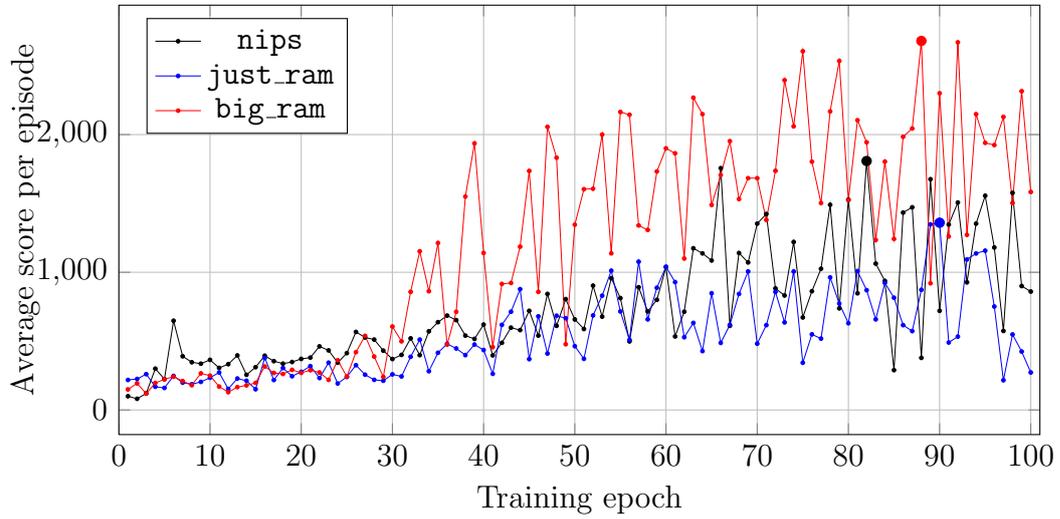

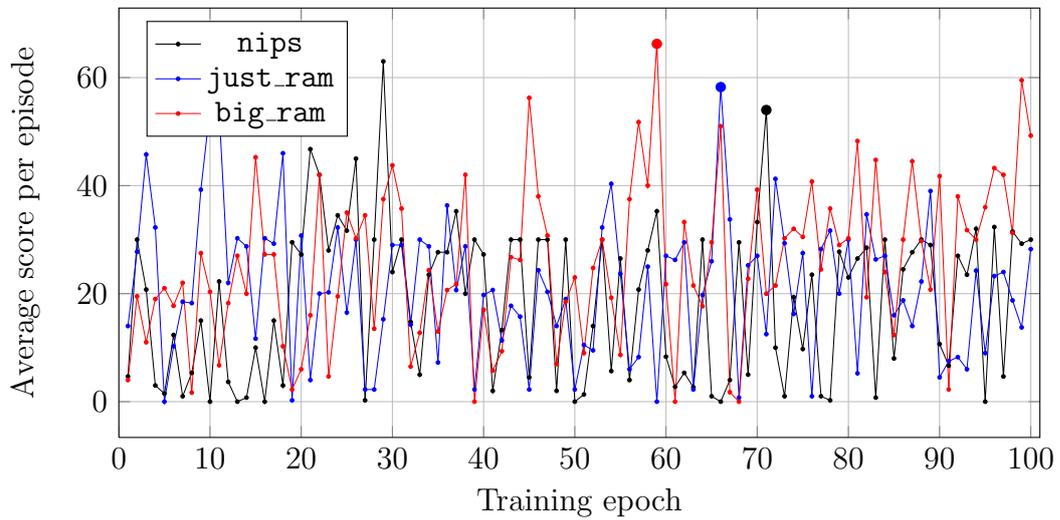
\begin{figure}[htpb]
\begin{center}
\hspace*{-1.5cm}\begin{tikzpicture}[scale=1]
\begin{axis}[
   xlabel={Training epoch},
   ylabel={Average score per episode},
   xmin=0,xmax=101,
   x=0.12cm,
   grid=major,
   mark options={solid, scale=0.35},
   legend entries={\texttt{nips}, \texttt{just\_ram}, \texttt{big\_ram}},
   legend style={legend pos=north west},
]
\addplot[mark=*, color=black, line width=0.25pt] table {experiments/bowling_nips.txt};
\addplot[mark=*, color=blue, line width=0.25pt] table {experiments/bowling_just_ram_100.txt};
\addplot[mark=*, color=red, line width=0.25pt] table {experiments/bowling_big_ram_100.txt};
\addplot[mark=*, color=black, only marks, mark size=5] table {
x y
71	54
};
\addplot[mark=*, color=blue, only marks, mark size=5] table {
x	y
66	58.25
};
\addplot[mark=*, color=red, only marks, mark size=5] table {
x	y
59 66.25
};
\end{axis}
\end{tikzpicture}
\end{center}
\caption{Training results for Bowling  for three plain models: \texttt{nips}, \texttt{just\_ram}, \texttt{big\_ram}.}
\label{fig:bowling-plain}
\end{figure}

Figures \ref{fig:breakout-plain},\ref{fig:seaquest-plain},\ref{fig:bowling-plain} show that there is a big variance of the results between epochs, especially in the RAM models. Because of that, to compare the models, we chose the results of the best epoch\footnote{For Breakout we tested networks with best training-time results. The test consisted of choosing other random seeds and performing $100\,000$ steps. For all  networks, including \texttt{nips}, we received results consistently lower by about $30\%$.}. We summarized these results in Table \ref{table:results-plain}, which also include the results of \texttt{ale\_ram}\footnote{The \texttt{ale\_ram}'s evaluation method differ -- the scores presented are the average over $30$ trials consisting of a long period of learning and then a long period of testing, nevertheless the results are much worse than of any DQN-based method presented here.}.
\begin{table}[H]
\begin{center}
\begin{tabularx}{0.7\textwidth}{ X c c c }
  \hline
  \rowcolor{green!40}
  &\ Breakout\ &\ Seaquest\ & \ Bowling\  \\
  \rowcolor{yellow!50}
  \texttt{nips} best & \textbf{213.14}  & $1808$  & $54.0$  \\
  \rowcolor{yellow!50}
  \texttt{just\_ram} best & $99.56$ & $1360$ & $58.25$  \\
  \rowcolor{yellow!50}
  \texttt{big\_ram} best & $178.0$ & \textbf{2680} & \textbf{66.25} \\
  \rowcolor{yellow!50}
  \texttt{ale\_ram} & $4.0$ & $593.7$ & $29.3$ \\
  \hline 
\end{tabularx}
\vspace{0.1in}
\caption{Table summarizing test results for basic methods.}
\label{table:results-plain}
\end{center}
\end{table}

In Breakout the best result for the \texttt{big\_ram} model is weaker than those obtained by the network \texttt{nips}. In Seaquest the best result obtained by the \texttt{big\_ram} network is better than the best result obtained by the network \texttt{nips}. In Bowling our methods give a slight improvement over the network \texttt{nips}, yet in all considered approaches the learning  as illustrated by Figure \ref{fig:bowling-plain} seem to be poor and the outcome in terms of gameplay is not very satisfactory. We decided to not include in this paper further experiments with Bowling and leave it as a topic of a further research.
\end{section}

%% file: regularization.tex
\begin{section}{Regularization}
Training a basic RAM-only network leads to high variance of the results (see the figures in the previous section) over epochs. This can be a sign of overfitting. To tackle this problem we've applied  {\em dropout} \cite{dropout}, a standard regularization technique for neural networks. 

Dropout is a simple, yet effective regularization method. It consists of ``turning off'' with probability $p$ each neuron in training, i.e. setting the output of the neuron to $0$, regardless of its input. In backpropagation, the parameters of switched off nodes are not updated. Then, during testing, all neurons are set to ``on'' -- they work as in the course of normal training, with the exception that each neuron's output is multiplied by $p$ to make up for the skewed training. The intuition behind the dropout method is that it forces each node to learn in absence of other nodes. The work \cite{dropout-analysis} shows an experimental evidence that the dropout method indeed reduces the variance of the learning process.

We've enabled dropout with probability of turning off a neuron $p=\frac{1}{2}$. This applies to all nodes, except output ones. We implemented dropout for two RAM-only networks: \texttt{just\_ram} and \texttt{big\_ram}. This method offers an  improvement for the \texttt{big\_ram} network leading to the best result for Seaquest in this paper. The best epoch results are presented in the Table \ref{table:results-regularization} and the intermediate training results for Seaquest are shown in Figure \ref{fig:seaquest-dropout}.
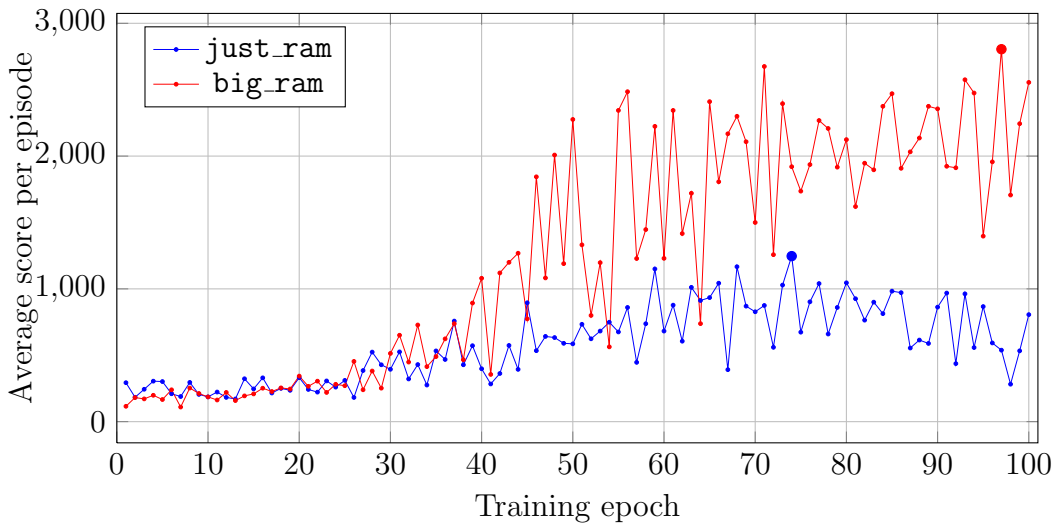
\begin{figure}[H]
\begin{center}
\hspace*{-1.5cm}\begin{tikzpicture}[scale=1]
\begin{axis}[
   xlabel={Training epoch},
   ylabel={Average score per episode},
   xmin=0,xmax=101,
   x=0.12cm,
   grid=major,
   mark options={solid, scale=0.35},
   legend entries={\texttt{just\_ram}, \texttt{big\_ram}},
   legend style={legend pos=north west},
]
\addplot[mark=*, color=blue, line width=0.25pt] table {experiments/seaquest_just_dropout.txt};
\addplot[mark=*, color=red, line width=0.25pt] table {experiments/seaquest_big_dropout.txt};
\addplot[mark=*, color=blue, only marks, mark size=5] table {
x	y
74	1246.66666667
};
\addplot[mark=*, color=red, only marks, mark size=5] table {
x	y
97	2805.0
};
\end{axis}
\end{tikzpicture}
\end{center}
\caption{Training results for Seaquest with dropout $p=0.5$ for models \texttt{just\_ram}, \texttt{big\_ram}. This figure suggests that indeed dropout reduces the variance of the learning process.}
\label{fig:seaquest-dropout}
\end{figure}

\begin{table}[h]
\begin{center}
\begin{tabularx}{0.8\textwidth}{ X c c }
  \hline
  \rowcolor{green!40}
  &\ Breakout\ &\ Seaquest\  \\
  \rowcolor{yellow!50}
  \texttt{just\_ram with dropout} best & $130.5$  & $1246.67$ \\
  \rowcolor{yellow!50}
  \texttt{big\_ram with dropout} best & $122.25$ & \textbf{2805} \\
  \hline 
\end{tabularx}
\vspace{0.1in}
\caption{Summary of test results for training which involves regularization with the dropout method with the parameter $p=0.5$.}
\label{table:results-regularization}
\end{center}
\end{table}

\vspace{-0.5in}
\section{Decreasing learning rate}
We also tried to reduce the variance of the learner through reduction of the learning rate from $0.0002$ to $0.0001$. 

The learning rate is a parameter of the algorithm {\em rmsprop}  that decides how parameters are changed in each step. Bigger learning rates correspond to moving faster in the parameter space, making learning faster, but more noisy.

We expected that the drastic changes in performance between consecutive epochs, as illustrated by Figures \ref{fig:breakout-plain} and \ref{fig:seaquest-plain}, may come from stepping over optimal values when taking too big steps. If it is the case, decreasing the step size should lead to slower learning combined with higher precision of finding minima of the loss function.

The results of these experiments can be found in Table \ref{table:rate}. Comparing to the training without regularization, scores improved only in the case of Breakout and the \texttt{just\_ram} network, but not by a big margin.

\begin{table}[H]
\begin{center}
\begin{tabularx}{0.7\textwidth}{ X c c  }
  \hline
  \rowcolor{green!40}
  &\ Breakout\ &\ Seaquest\ \\
  \rowcolor{yellow!50}
  \texttt{just\_ram} best & $137.67$ & $1233.33$  \\
  \rowcolor{yellow!50}
  \texttt{big\_ram} best  & $112.14$ & $2675$ \\
  \hline 
\end{tabularx}
\vspace{0.1in}
\caption{Summary of test results for modified learning rate.}
\label{table:rate}
\end{center}
\end{table}
\end{section}

%% file: frameskip.tex
\begin{section}{Frame skip}
Atari 2600 was designed to use an analog TV as the output device with 60 new frames appearing on the screen every second. To simplify the search space we impose a rule that one action is repeated over a fixed number of frames. This fixed number is called the {\em frame skip}. The standard frame skip used in \cite{mnih-atari-2013} is 4. For this frame skip the agent makes a decision about the next move every $4\cdot \frac{1}{60} = \frac{1}{15}$ of a second. Once the decision is made, then it remains unchanged during the next $4$ frames.

Low frame skip allows the network to learn strategies based on a super-human reflex. High frame skip will limit the number of strategies, hence learning may be faster and more successful.

In the benchmark agent \texttt{nips}, trained with the frame skip $4$, \textit{all} $4$ frames are used for training along with the sum of the rewards coming after them. This is dictated by the fact that due to hardware limitations, Atari games sometimes "blink", i.e. show some objects only every few frames. For example, in the game Space Invaders, if an enemy spaceship shoots in the direction of the player, then shots can be seen on the screen only every second frame and an agent who sees only the frames of the wrong parity would have no access to a critical part of the game information.

In the case of learning from memory we are not aware of any critical loses of information when intermediate RAM states are ignored. Hence in our models we only passed to the network the RAM state corresponding to the last frame corresponding to a given action\footnote{We also tried to pass all the RAM states as a ($128*\texttt{FRAME SKIP}$)-dimensional vector, but this didn't lead to an improved performance.}.

The work \cite{frameskip} suggests that choosing the right frame skip can have a big influence on the performance of learning algorithms (see also \cite{microsoft_res}). Figure \ref{fig:seaquest-fs} and Table \ref{table:results-frameskip} show a significant improvement of the performance of the \texttt{just\_ram} model in the case of Seaquest.
Quite surprisingly, the variance of results appeared to be much lower for higher \texttt{FRAME SKIP}.

As noticed in \cite{frameskip}, in the case of Breakout high frame skips, such as $\texttt{FRAME SKIP} = 30$, lead to a disastrous performance. Hence we tested only lower  $\texttt{FRAME SKIP}$ and for $\texttt{FRAME SKIP} = 8$ we received results slightly weaker than those with $\texttt{FRAME SKIP} = 4$.

\begin{table}[h]
\begin{center}
\begin{tabularx}{0.9\textwidth}{ X c c }
  \hline
  \rowcolor{green!40} &\ Breakout\ &\ Seaquest\  \\
  \rowcolor{yellow!50}
  \texttt{just\_ram} with \texttt{FRAME SKIP} $8$ best & $82.87$  & $2064.44$ \\
  \rowcolor{yellow!50}
  \texttt{just\_ram} with \texttt{FRAME SKIP} $30$ best & -  & $2093.24$ \\
  \rowcolor{yellow!50}
  \texttt{big\_ram} with \texttt{FRAME SKIP} $8$ best & $102.64$ & $2284.44$ \\
  \rowcolor{yellow!50}
  \texttt{big\_ram} with \texttt{FRAME SKIP} $30$ best & - & $2043.68$ \\
  \hline 
\end{tabularx}
\vspace{0.1in}
\caption{Table summarizing test results for training which involves higher \texttt{FRAME SKIP} value. For Breakout \texttt{FRAME SKIP}$
=30$ does not seem to be a suitable choice.}
\label{table:results-frameskip}
\end{center}
\end{table}
\begin{figure}[h]
\begin{center}
\hspace*{-1.5cm}\begin{tikzpicture}[scale=1]
\begin{axis}[
   xlabel={Training epoch},
   ylabel={Average score per episode},
   xmin=0,xmax=101,
   x=0.12cm,
   grid=major,
   mark options={solid, scale=0.35},
   legend entries={$\texttt{FRAME SKIP} = 8$, $\texttt{FRAME SKIP} = 30$},
   legend style={legend pos=north west},
]
\addplot[mark=*, color=blue, line width=0.25pt] table {experiments/seaquest_just_fs8.txt};
\addplot[mark=*, color=red, line width=0.25pt] table {experiments/seaquest_just_fs30.txt};
\addplot[mark=*, color=blue, only marks, mark size=5] table {
x	y
85	2064.44444444
};
\addplot[mark=*, color=red, only marks, mark size=5] table {
x	y
100	2093.24324324
};
\end{axis}
\end{tikzpicture}
\end{center}
\caption{Training results for Seaquest with 
$\texttt{FRAME SKIP} = 8$ and  $\texttt{FRAME SKIP} = 30$ for the model \texttt{just\_ram}.}
\label{fig:seaquest-fs}
\end{figure}
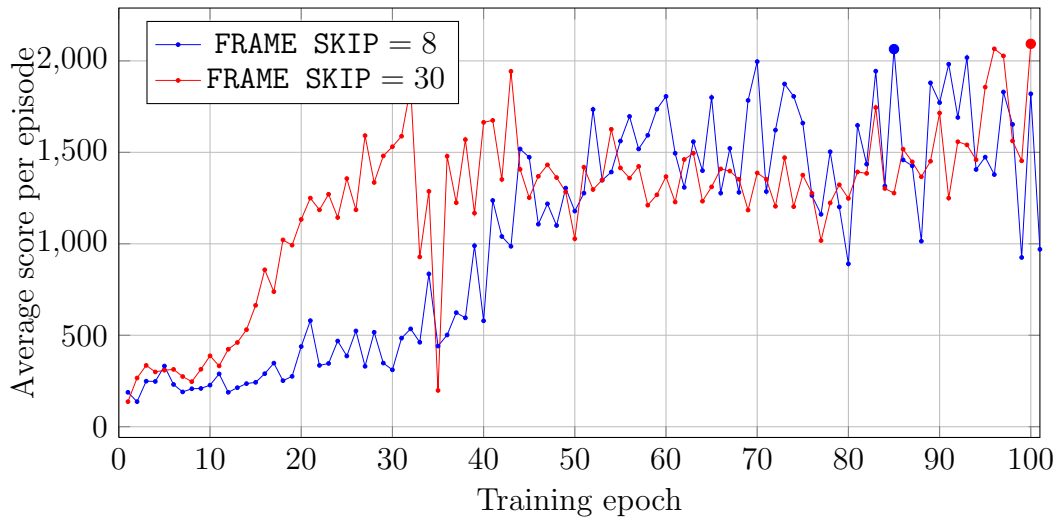
\end{section}

%% file: other.tex
\begin{section}{Mixing screen and memory}

One of the hopes of future work is to integrate the information from the RAM and information from the screen in order to train an ultimate Atari 2600 agent. In this work we made some first steps towards this goal. We consider two mixed network architectures. The first one is \texttt{mixed\_ram}, where we just concatenate the output of the last hidden layer of the convolutional network with the RAM input and then in the last layer apply a linear transformation without any following non-linearity.

\RestyleAlgo{ruled}
\begin{algorithm}[H]
\caption{\texttt{mixed\_ram}(outputDim)}

\SetAlgoVlined
\LinesNumbered

\KwIn{RAM,screen} 
\KwOut{A vector of length outputDim}
\vspace{0.05cm}

$conv1 \leftarrow Conv2DLayer(screen,rectify)$

$conv2 \leftarrow Conv2DLayer(conv1,rectify)$

$hidden \leftarrow DenseLayer(conv2,256,rectify)$

$concat \leftarrow ConcatLayer(hidden,RAM)$

$output \leftarrow DenseLayer(concat,outputDim,no\; activation)$

\Return{$output$}
\end{algorithm}

The other architecture is a deeper version of \texttt{mixed\_ram}. We allow more dense layers which are applied in a more sophisticated way as described below. 

\RestyleAlgo{ruled}
\begin{algorithm}[H]
\caption{\texttt{big\_mixed\_ram}(outputDim)}

\SetAlgoVlined
\LinesNumbered

\KwIn{RAM,screen} 
\KwOut{A vector of length outputDim}
\vspace{0.05cm}

$conv1 \leftarrow Conv2DLayer(screen,rectify)$

$conv2 \leftarrow Conv2DLayer(conv1,rectify)$

$hidden1 \leftarrow DenseLayer(conv2,256,rectify)$

$hidden2 \leftarrow DenseLayer(RAM,128,rectify)$

$hidden3 \leftarrow DenseLayer(hidden2,128,rectify)$

$concat \leftarrow ConcatLayer(hidden1,hidden3)$

$hidden4 \leftarrow DenseLayer(concat,256,rectify)$

$output \leftarrow DenseLayer(hidden4,outputDim,no\; activation)$

\Return{$output$}
\end{algorithm}
The obtained results presented in Table \ref{table:mixed} are reasonable, but not particularly impressive. In particular we did not notice any improvement over the benchmark \texttt{nips} network, which is embedded into both mixed architectures. This suggests that in the  \texttt{mixed\_ram} and  \texttt{big\_mixed\_ram} models the additional information from the memory is not used in a productive way. 
\begin{table}[H]
\begin{center}
\begin{tabularx}{0.7\textwidth}{ X c c  }
  \hline
  \rowcolor{green!40}
  &\ Breakout\ &\ Seaquest\ \\
  \rowcolor{yellow!50}
  \texttt{mixed\_ram} best & $143.67$ & $488.57$ \\
  \rowcolor{yellow!50}
  \texttt{big\_mixed\_ram} best  & $67.56$ & $1700$ \\
  \hline 
\end{tabularx}
\vspace{0.1in}
\caption{Table summarizing test results for methods involving information from the screen and from the memory.}
\label{table:mixed}
\end{center}
\end{table}
\end{section}

\begin{section}{RAM visualization}

We visualized the first layers of the neural networks in an attempt to understand how they work. Each column in Figure \ref{fig:ram-vis} corresponds to one of $128$ nodes in the first layer of the trained \texttt{big\_ram} network and each row corresponds to one of $128$ memory cells. The color of a cell in a given column describes whether the high value in this RAM cell negatively (blue) or positively (red) influence the activation level for that neuron.  Figure \ref{fig:ram-vis} suggests that the RAM cells with numbers $95-105$ in Breakout and $90-105$ in Seaquest are important for the gameplay -- the behavior of \texttt{big\_ram} networks depend to the the high extent on the state of these cells.
\begin{figure}[!h]
\centering
\subfigure[\texttt{big\_ram} in Breakout]{
\includegraphics[scale=0.3]{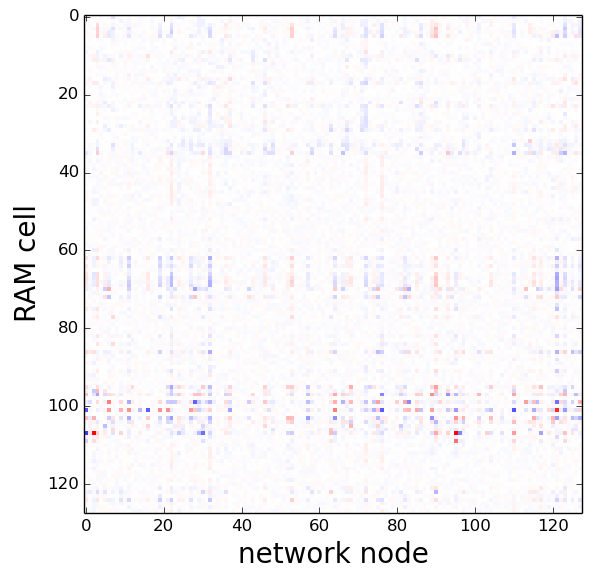}
}
\hspace{15pt}
\subfigure[\texttt{big\_ram} in Seaquest]{
\includegraphics[scale=0.299]{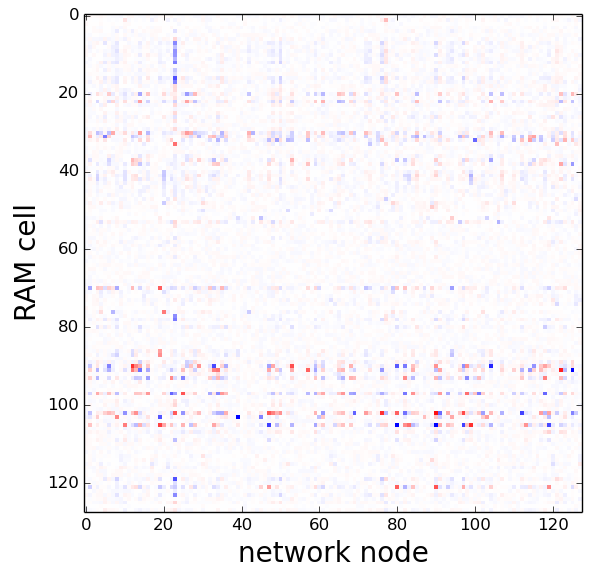}
}
\caption{Visualization of the parameters of the first layer of the trained Q-networks.}
\label{fig:ram-vis}
\end{figure}
\end{section}

%% file: conclusion.tex
\begin{section}{Conclusions}
We trained a number of neural networks capable of playing Atari 2600 games: Bowling, Breakout and Seaquest. The novel aspect of this work is that the networks use information stored in the memory of the console.  In all games the RAM agents are on a par with the screen-only agent \texttt{nips}. The RAM agents trained using methods described in this work were unaware of more abstract features of the games, such as counters controlling amount of oxygen or the number of divers in Seaquest.

In the case of Seaquest, even a simple \texttt{just\_ram} architecture with an appropriately chosen \texttt{FRAME SKIP} parameter as well as the \texttt{big\_ram} agent with standard parameters, performs better than the benchmark \texttt{nips} agent. 
In the case of Breakout, the performance is below the screen-only agent \texttt{nips}, but still reasonable. In the case of of Bowling  methods presented in \cite{mnih-atari-2013} as well as those in this paper are not very helpful -- the agents play at a rudimentary level.
\end{section}

\begin{section}{Future work}
\begin{subsection}{Games with more {\em refined} logic}
Since in the case of Seaquest the performance of RAM-only networks is quite good, a natural next target would be games such as Pacman or Space Invaders, which similarly to Seaquest offer interesting tactical challenges.
\end{subsection}

\begin{subsection}{More sophisticated architecture and better hyperparameters}
The recent papers \cite{mnih-dqn-2015,van2015deep,liang2015state,duelling} introduce more sophisticated ideas to improve deep Q-networks. We would like to see whether these improvements also apply to the RAM models.

It would be also interesting to tune hyperparameters in a way which would specifically address the needs of RAM-based neural networks. In particular we are interested in:
\begin{itemize}
\item better understanding what the deep Q-network learns about specific memory cells; can one identify critical cells in the memory?
\item improving stability of learning and reducing variance and overfitting,
\item more effective joining of information from the screen and from the memory,
\item trying more complex, deeper architectures for RAM.
\end{itemize}
\end{subsection}

\begin{subsection}{Recurrent neural networks and patterns of memory usage}
Reading the RAM state while running the deep Q-learning algorithm gives us an access to a practically unlimited stream of Atari 2600 memory states. We can use this stream to build a recurrent neural network which takes into account previous RAM states.

In our view it would also be interesting to train an {\em autoencoder}. It may help to identify RAM patterns used by Atari 2600 programmers and to find better initial parameters of the neural network \cite{autoencoders}.
\end{subsection}

\begin{subsection}{Patterns as finite state machines}
The work of D.~Angluin \cite{angluin} introduced the concept of learning the structure of a finite state machine through queries and counter-examples. A game for Atari 2600 can be identified with a finite state machine which takes as input the memory state and action and outputs another memory state. We are interested in devising a neural network which would learn the structure of this finite state machine.  
The successive layers of the network would learn about sub-automata responsible for specific memory cells and later layers would join the automata into an automaton which would act on the whole memory state.
\end{subsection}
\end{section}

%% file: acknowledge.tex
\begin{section}{Acknowledgements}
This research was carried out with the support of grant GG63-11 awarded by the Interdisciplinary Centre for Mathematical and Computational Modelling (ICM) University of Warsaw.

We would like to express our thanks to Marc G. Bellemare for suggesting this research topic.
\end{section}

%% file: append.tex
\appendix
\begin{section}{Parameters} \label{app:AppendixA}
The list of hyperparameters and their descriptions. Most of the descriptions come from \cite{mnih-dqn-2015}.
    
\begin{table}[H]
\begin{tabularx}{\textwidth}{@{} l c Y @{}}
\toprule
hyperparameter & value & description \\
\midrule
minibatch size & $32$ & Number of training cases over which each stochastic gradient descent (SGD) update is computed. \\\addlinespace
replay memory size & $100\,000$ & SGD updates are randomply sampled from this number of most recent frames. \\\addlinespace
phi length & $4$ & The number of most recent frames experienced by the agent that are given as input to the Q network in case of the networks that accept screen as input. \\\addlinespace
update rule & \texttt{rmsprop} &  Name of the algorithm optimizing the neural network's objective function   \\ \addlinespace
learning rate &  $0.0002$  & The learning rate for rmsprop\\ \addlinespace
discount & $0.95$ & Discount factor $\gamma$ used in the Q-learning update. Measures how much less do we value our expectation of the value of the state in comparison to observed reward.\\\addlinespace
epsilon start & $1.0$ & The probability ($\varepsilon$) of choosing a random action at the beginning of the training. \\\addlinespace
epsilon decay & $1000000$ & Number of frames over which the $\varepsilon$ is faded out to its final value. \\\addlinespace
epsilon min & $0.1$ & The final value of $\varepsilon$, the probability of choosing a random action. \\\addlinespace
replay start size & $100$ & The number of frames the learner does just the random actions to populate the replay memory.\\
\bottomrule
\end{tabularx}
\caption{Parameters} 
\label{table:param}
\end{table}
\end{section}